\UseRawInputEncoding

\documentclass[letterpaper, 10 pt, conference]{ieeeconf}  

\IEEEoverridecommandlockouts                              

\usepackage{cite}
\usepackage{amsmath,amssymb,amsfonts}
\usepackage{algorithmic}
\usepackage{graphicx}
\usepackage{bbding}
\usepackage{textcomp}
\usepackage{bm}
\usepackage{color}
\usepackage{float}
\usepackage{subfigure}
\usepackage{stfloats}
\usepackage{diagbox}
\usepackage{multirow}
\usepackage{harpoon}
\usepackage{booktabs}

\newcommand\blfootnote[1]{%
	\begingroup 
	\renewcommand\thefootnote{}\footnote{#1}%
	\addtocounter{footnote}{-1}%
	\endgroup 
}

\title{\LARGE \bf
Contrastive Label Disambiguation for Self-Supervised Terrain Traversability Learning in Off-Road Environments
}

\author{Hanzhang Xue$^{1, 2}$,  Xiaochang Hu$^{1}$, Rui Xie$^{3}$, Hao Fu$^{1}$, Liang Xiao$^{2}$, Yiming Nie$^{2}$, Bin Dai$^{2}$ 
}

\begin{document}

\twocolumn[{%
	\renewcommand\twocolumn[1][]{#1}%
	\maketitle
	\begin{figure}[H]
		\vspace{-0.6cm} 
		\hsize=\textwidth
		\subfigure {
			\centering
			\includegraphics[width=0.315\textwidth]{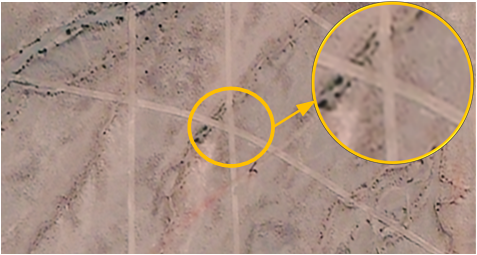}
		}
		\subfigure {
			\centering
			\includegraphics[width=0.315\textwidth]{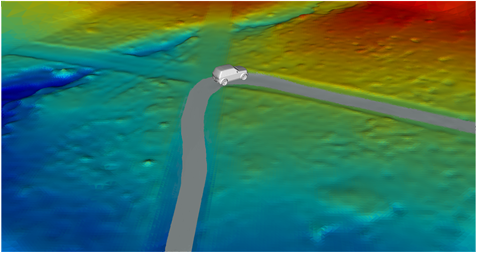}
		}  
		\subfigure {
			\centering
			\includegraphics[width=0.315\textwidth]{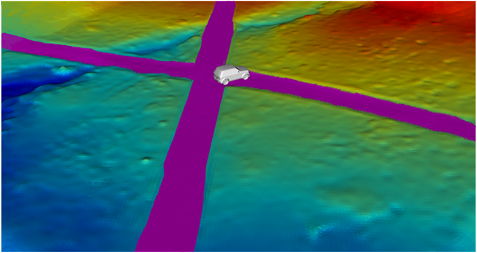}
		}
		\caption{The left figure is the satellite image corresponding to our Gobi Desert driving dataset. The middle figure shows the training terrain samples automatically annotated from actual driving experiences. The background is the terrain model constructed in real time, different colors indicate different elevations. Gray are regions traversed by the UGV, these regions are labeled as positive samples, and the remaining regions are unlabeled samples. The right figure demonstrates the final traversability analysis results, magenta are traversable regions learned from driving experiences of the UGV.}
		\label{fig_intro}
		\vspace{-0.3cm} 
	\end{figure}
}]

\blfootnote{*This work was supported by the National Natural Science Foundation of China under No.61790565 and No.61803380.}
\blfootnote{$^{1}$Hanzhang Xue, Xiaochang Hu, and Hao Fu are with the College of Intelligence Science and Technology, National University of Defense Technology, Changsha, 410073, China {\tt\small xuehanzhang13@nudt.edu.cn, huxiaochang17@nudt.edu.cn, fuhao@nudt.edu.cn}.}
\blfootnote{$^{2}$Hanzhang Xue, Liang Xiao, Yiming Nie, and Bin Dai are with the Unmanned Systems Technology Research Center, Defense Innovation Institute, Beijing, 100071, China {\tt\small xiaoliang@nudt.edu.cn, 13370154812@189.cn, ibindai@163.com}.}%
\blfootnote{$^{3}$Rui Xie is with the Key Laboratory of Machine Perception (MOE), Peking University, Beijing, 100871, China {\tt\small xierui1124@stu.pku.edu.cn}.}

%
%
%
\begin{abstract}
Discriminating the traversability of terrains is a crucial task for autonomous driving in off-road environments. However, it is challenging due to the diverse, ambiguous, and platform-specific nature of off-road traversability. In this paper, we propose a novel self-supervised terrain traversability learning framework, utilizing a contrastive label disambiguation mechanism. Firstly, weakly labeled training samples with pseudo labels are automatically generated by projecting actual driving experiences onto the terrain models constructed in real time. Subsequently, a prototype-based contrastive representation learning method is designed to learn distinguishable embeddings, facilitating the self-supervised updating of those pseudo labels. As the iterative interaction between representation learning and pseudo label updating, the ambiguities in those pseudo labels are gradually eliminated, enabling the learning of platform-specific and task-specific traversability without any human-provided annotations. Experimental results on the RELLIS-3D dataset and our Gobi Desert driving dataset demonstrate the effectiveness of the proposed method.

\end{abstract}

\section{INTRODUCTION}
For autonomous driving, understanding the traversability of surrounding environments is one of the most fundamental and critical tasks. The majority of existing related works primarily concentrates on structured environments where the traversable regions are explicitly defined, and treat the traversability analysis as a binary classification task. Closely related tasks include road detection \cite{Gu2021}, ground segmentation \cite{Lee2022}, or free-space detection \cite{Min2022}. However, most of these approaches may not work well in complex off-road environments. There are two main reasons: Firstly, there is a high degree of similarity between traversable and non-traversable regions in some off-road environments; Secondly, countless terrain types and irregular terrain shapes in off-road environments present intricate possibilities for traversability, and it is challenging to analyze them with a unified rule.
 
Recently, several researchers \cite{Maturana2017, Shaban2021} have attempted to employ supervised semantic segmentation approaches to obtain semantic information for each region in off-road environments, and to analyze traversability by establishing mapping relationships between different semantic categories and traversability. Although these methods achieved impressive results on some specific datasets, they are difficult to adapt to previously unseen environments since the inherent ambiguity of the traversability in off-road environments. On the one hand, it is challenging to define semantic categories and their corresponding traversability in diverse off-road environments without ambiguity. On the other hand, traversability itself is also platform-related and task-related, different platforms or autonomous tasks may yield different traversability results. Furthermore, these supervised learning approaches also require exhausting human labor for manual annotation of training samples. It is unaffordable and unsustainable for re-annotating a tremendous amount of data each time a new environment is encountered.

Bearing the purpose of rapidly learning platform-specific and task-specific traversability in new off-road environments without any human-provided annotations, we shift from directly define traversability or semantic categories in off-road environments to understand traversability by learning from demonstration. When an unmanned ground vehicle (UGV) encounters an unknown off-road environment or a new autonomous task, a large amount of weakly labeled data can be automatically generated by simply driving the UGV for a short distance with the assistance of a human driver (shown as the middle figure in Fig. \ref{fig_intro}). As derived from actual driving experiences, these weakly labeled data are platform-specific and task-specific. They consist of scarce positive samples (actually traversed regions) and numerous unlabeled samples, which can be employed as training data in the problem of self-supervised traversability learning. Recently, although a few approaches such as positive-unlabeled learning \cite{Suger2015} or anomaly detection \cite{Wellhausen2020} have been applied to address this problem, they possess limited ability to discriminate traversability in off-road environments with high similarity. Furthermore, common forms of input data for this problem include images \cite{Schmid2022} or single-frame LiDAR scan \cite{Bae2022}. Images are sensitive to illumination changes, and single-frame LiDAR scan is sparse and prone to noises. Consequently, neither can provide a stable and robust representation of off-road environments, directly affecting the stability of traversability learning.

To address these challenges, we propose a novel self-supervised terrain traversability learning framework. After generating stable, complete, and accurate terrain models in real-time using our previous work \cite{Xue2023}, automatic data annotation is conducted in those constructed terrain models based on actual driving experiences. Those actually traversed regions are assigned determined positive labels, while the remaining regions are assigned candidate pseudo labels. Inspired by the impressive progress of partial label learning (PLL) \cite{Wang2022}, a prototype-based contrastive representation learning method with the aid of a local window based transformer encoder is designed to learn distinguishable embeddings for updating those candidate pseudo labels, and the refined pseudo labels in turn facilitate representation learning. As the iterative interaction between representation learning and label updating, the ambiguities associated with those pseudo labels are gradually eliminated, enabling the learning of specific traversability in off-road environments. This learning process can be referred to as contrastive label disambiguation.  

To demonstrate the effectiveness of the proposed method, we conduct experiments on both the publicly available RELLIS-3D dataset \cite{Jiang2021} and a Gobi Desert driving dataset collected by our own UGV. Experimental results show that the proposed method can learn specific traversability from human-selected driving routes in a self-supervised manner.

The rest of this paper is organized as follows. Section \ref{section:related_works} discusses some related works. Section \ref{Section: our_method} provides detailed information about the proposed method. Experimental results on both the RELLIS-3D dataset and our Gobi Desert driving dataset are presented in Section \ref{Section: experiment}. Finally, Section \ref{Section: Conclusion} summarizes the conclusions.

\section{RELATED WORK} \label{section:related_works}
\begin{figure*}[t]
	\centering
	\includegraphics[scale=0.198]{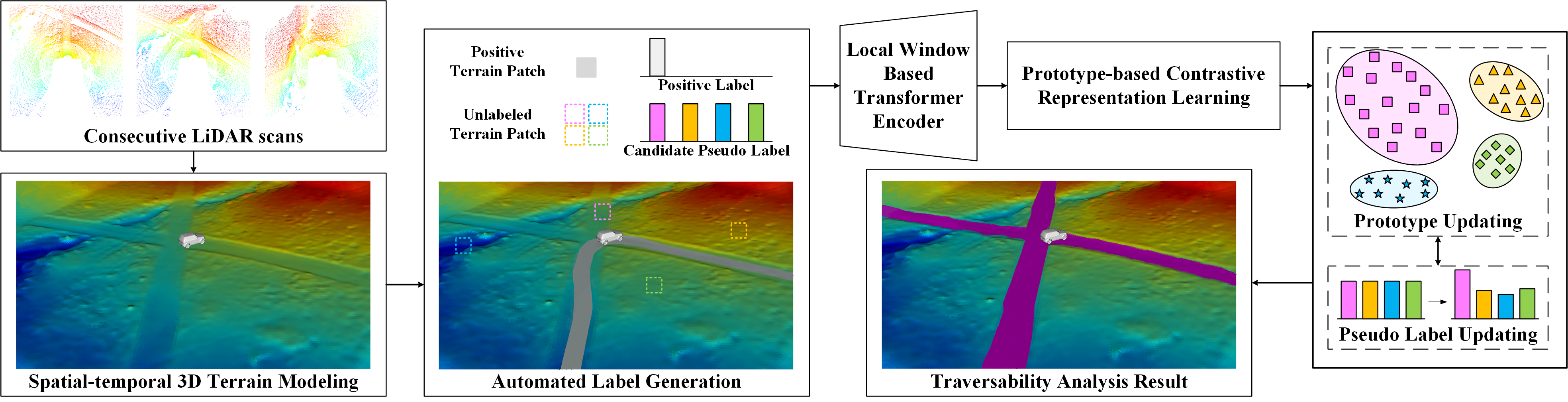} \\
	\caption{The pipeline for the proposed self-supervised traversability learning framework.}
	\label{fig_Framework}
	\vspace{-0.6cm} 
\end{figure*}

\begin{figure*}[bp]
	\begin{minipage}{0.33\textwidth}
		\vspace{-0.4cm}
		\centering
		\includegraphics[scale=0.33]{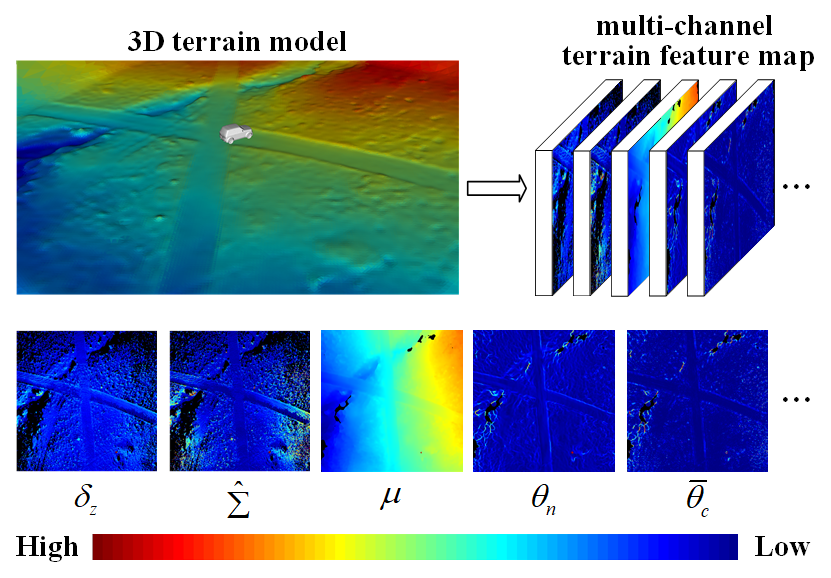}
		\caption{An illustration of the multi-channel terrain feature map. Some feature channels are visualized in the bottom. The magnitude of $\delta_z$, $\hat{\Sigma}$, $\mu$, $\theta_n$, and $\bar{\theta}_c$ for each cell is visualized by different colors.}
		\label{fig_feature_map}
	\end{minipage}\hfill
	\begin{minipage}{0.64\textwidth}
		\vspace{-0.4cm} 
		\centering
		\includegraphics[scale=0.34]{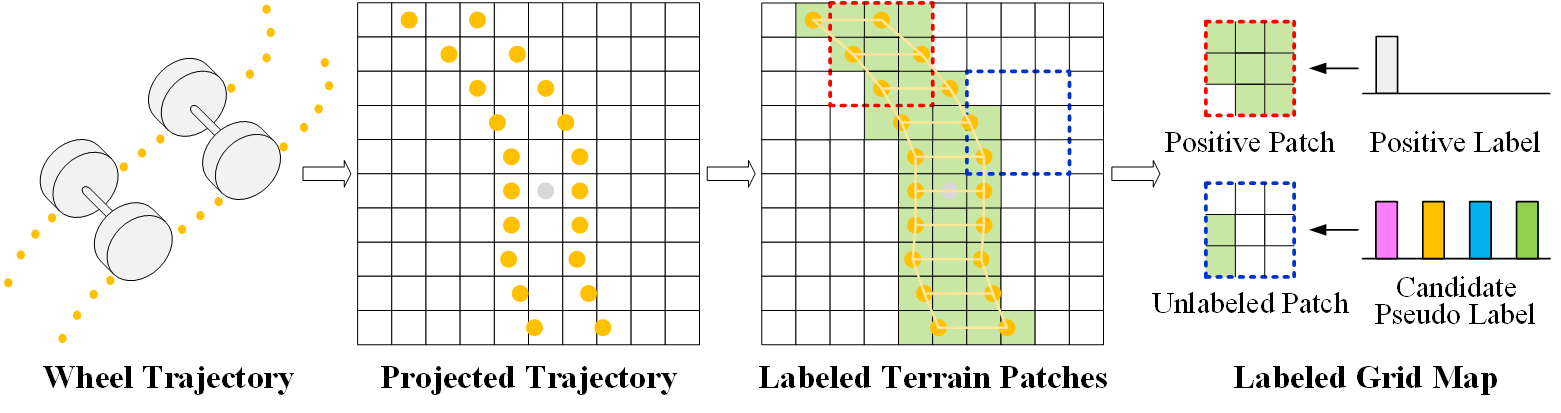}
		\caption{The process of the automated label generation. Past and future vehicle trajectories (denoted as yellow circles) are first projected into the current feature map. Four projected points from each trajectory are then connected to form a quadrilateral, those cells lying within the quadrilateral are annotated as positive cells (colored by green). The red square indicates a positive terrain patch with a determined positive label, while the blue square is an unlabeled terrain patch with a candidate pseudo label. }
		\label{fig_vehicle_trajectory}
	\end{minipage}\hfill	
\end{figure*}

Traversability analysis plays a crucial role in autonomous driving and has garnered significant attention in recent years. In the existing literature, most approaches treat traversability analysis as a binary classification task. One common method projects point cloud or RGB images onto a 2D Bird's Eye View (BEV) grid map and extracts geometric features \cite{Douillard2010} or appearance features \cite{Lu2014} for traversability classification. Another kind of approaches determines traversability by estimating terrain models of local environments, such as Gaussian process regression \cite{Chen2014}, Bayesian generalized kernel inference \cite{Shan2018}, and B-spline surface \cite{Rodrigues2020}. With the rise of deep learning, convolutional neural networks (CNNs) have also been utilized for end-to-end traversable region detection by using RGB images \cite{Min2022}, point cloud \cite{Paigwar2020}, or a combination of both \cite{Gu2021} as input. Although these binary classification approaches work well in structured environments, their suitability for complex off-road environments is limited. 

It is more suitable to adopt semantic mapping for traversability analysis in off-road environments, which allows for distinction of semantic information among different regions. Some methods \cite{Maturana2017, Holder2016} perform fine-grained semantic mapping to assign a detailed semantic label (such as dirt roads, grass, bushes, etc.) to each region by using semantic segmentation networks. The traversability can be further analyzed through mapping relationships between semantic categories and traversability. Some other works \cite{Schilling2017, Shaban2021, Guan2022} argue that fine-grained semantic information is not necessary for autonomous navigation and propose solutions based on coarse-grained semantic mapping. In these approaches, different regions are segmented by traversability levels. Although these semantic mapping approaches achieve good results in some specific off-road environments, they face some intractable challenges. Firstly, it is difficult to define uniform and unambiguous semantic categories or semantic-traversability mapping relationships that are suitable for all off-road environments. Additionally, these methods heavily rely on supervised labels and the burden of manually re-annotating pixel-level or grid-level semantic labels each time a new environment is encountered proves to be impractical, thus limiting the practical application of these approaches.

Recently, there has been a growing interest in self-supervised learning methods for traversability analysis. The physical experiences of the UGV, rather than human-provided annotations, are used to automatically label the training data. These methods can be divided into two categories. The first type utilizes on-board proprioceptive sensors to measure signals that directly reflect information about terrain traversability. Commonly used sensors include the Inertial Measurement Unit (IMU) \cite{Castro2023, Otsu2016, Seo2023}, force-torque sensors \cite{Wellhausen2019}, or acoustic sensors \cite{Zurn2021}. Then, traversability-related signals are used to annotate data from exteroceptive sensors (such as images or point cloud), and the generated weakly labeled data is employed as training samples for self-supervised learning traversability classification \cite{Wellhausen2019, Zurn2021, Otsu2016, Seo2023} or regression \cite{Castro2023, Wellhausen2019, Seo2023}. The second category of methods obtain labeled data directly from the driving experiences of the UGV. Vehicle trajectories are projected into image space or point cloud space, successful and failed traverses provide positive and negative traversability labels, respectively. Some works \cite{Suger2015, Schmid2022, Navarro2015} utilize only positive samples annotated by the footprints of the UGV for learning traversability in a positive-unlabeled learning manner. Part of works try to enhance performance by incorporating negative samples, for example, using LiDAR-based obstacle detection algorithms to annotate obstacle regions as negative samples \cite{Gao2019, Barnes2017, Lee2021}. However, these approaches can not distinguish those non-traversable regions that are not obstacles. Additionally, Chavez-Garcia et al. \cite{Garcia2018} employ a simulation system for self-learning traversability estimation, labeling regions where the UGV gets stuck as negative samples. Bae et al. \cite{Bae2022} introduce a small amount of manually labeled support data to provide negative samples, resulting in better performance with reduced labor costs.

\section{THE PROPOSED APPROACH} \label{Section: our_method}
In this paper, a self-supervised traversability learning framework is proposed with treating this problem as a contrastive label disambiguation task. The pipeline of the proposed framework is illustrated in Fig. \ref{fig_Framework}, consisting of four core modules: a spatial-temporal 3D terrain modeling module, an automated label generation module, a local window based transformer encoder, and a prototype-based contrastive representation learning module.

\subsection{Spatial-temporal 3D Terrain Modeling}
\label{Section: Terrain}
To provide a stable, complete, and objective representation of off-road environments and overcome the limitations of sparse and noise-prone point cloud data, a spatial-temporal terrain modeling approach proposed in our previous work \cite{Xue2023} is applied to generate dense 3D terrain models in real-time. In this approach, a normal distributions transform (NDT) mapping technique is first utilized to recursively fuse information from consecutive LiDAR scans into a global grid map. The elevation of each observed grid cell is modeled as a normal distribution $\mathcal{N} (\hat{\mu}, \hat{\Sigma})$. Subsequently, a bilateral filtering-aided Bayesian generalized kernel (BGK) inference approach is employed to infer a predicted elevation distribution $\mathcal{N} \left(\mu, \Sigma\right)$ for each grid cell, thus producing a dense and stable elevation map. Furthermore, various terrain features can be calculated by considering the geometric connectivity properties between adjacent grid cells. 

After constructing the 3D terrain models, a multi-channel terrain feature map $\bm{F}$ can be generated by projecting various terrain features onto the 2D grid map (as shown in Fig. \ref{fig_feature_map}). The features contained within each grid cell include: (1) The mean $\hat{\mu}$ and variance $\hat{\Sigma}$ of the observed elevation distribution; (2) The mean $\mu$ and variance $\Sigma$ of the predicted elevation distribution; (3) The maximum-minimum observed elevation difference $\delta_z$; (4) The normal angle $\theta_n$; (5) The average concavity angle $\bar{\theta}_c$ \cite{Moosmann2009} between the 4-neighboring grid cells.

\subsection{Automated Label Generation}
\label{Section: Label}
To provide training samples without any human-provided annotations, vehicle trajectories are utilized to annotate terrain patches in the proposed method. The process of the automated label generation is illustrated in Fig. \ref{fig_vehicle_trajectory}. 

A vehicle trajectory is defined as positions of four contact points $\bm{P} = \left\{P_{lf}, P_{rf}, P_{lr}, P_{rr}\right\}$ between four vehicle wheels and the ground plane. The past or future vehicle trajectory at timestamp $\tau$ can be transformed into the local body coordinate system $\left\{B\right\}$ at current timestamp $t$ by: 
\begin{equation}
\bm{P}^B_{\tau, t} = \left(\bm{T}^{WB}_t\right)^{-1} \cdot \bm{T}^{WB}_{\tau} \cdot \bm{P}^B \,,
\end{equation}
where $\bm{P}^B_{\tau, t}$ represents the transformed vehicle trajectory. $\bm{P}^B$ denotes the local position of the contact points $\bm{P}$, which can be measured by a simple calibration process. $\bm{T}^{WB}_{\tau}$ denotes the transformation matrix from the global coordinate system $\left\{W\right\}$ into the local body coordinate system $\left\{B\right\}$ at timestamp $\tau$, and it is estimated by using an online pose estimation module proposed in our previous work \cite{Xue2019}. 

The transformed vehicle trajectory set $\left\{\bm{P}^B_{\tau, t}\right\}_{\tau \in \left[t_p, t_f\right]}$ ($\left[t_p, t_f\right]$ denotes valid time interval) is projected onto the terrain feature map $\bm{F}_t$ generated at current timestamp $t$. Four projected points from each vehicle trajectory are connected to form a quadrilateral in $\bm{F}_t$. As shown in Fig. \ref{fig_vehicle_trajectory}, those grid cells lying within the quadrilateral are considered as actually traversed regions and are annotated as positive cells $G_p$, while the remaining grid cells are unlabeled cells $G_u$. Then, weakly annotated terrain patches can be extracted. Each terrain patch $\bm{x}_i$ consists of $M \times M$ grid cells ($M$ is an odd number), and its pseudo label $\bm{y}_i$ is determined by the annotation of its central grid cell $G_{i,c}$. $\bm{y}_i$ is represented as a one-hot encoded form:
\begin{equation}
\begin{aligned}
\bm{y}_i = & \left\{ 
\begin{matrix}
	\left[1 \quad 0 \quad 0 \quad \cdots \quad 0 \right] \quad \left(G_{i,c} = G_p \right) \,, \\
	\left[1 \quad 1 \quad 1 \quad \cdots \quad 1 \right] \quad \left(G_{i,c} = G_u \right) \,,
\end{matrix} \right. \\[-4mm]
& \quad \underbrace{\phantom{1 \quad 1 \quad 1 \quad \cdots \quad 1}}_K
\end{aligned}
\end{equation}
where $K$ is the total number of traversability categories.

\subsection{Local Window Based Transformer Encoder}
\begin{figure}[t]
	\centering
	\includegraphics[scale=0.39]{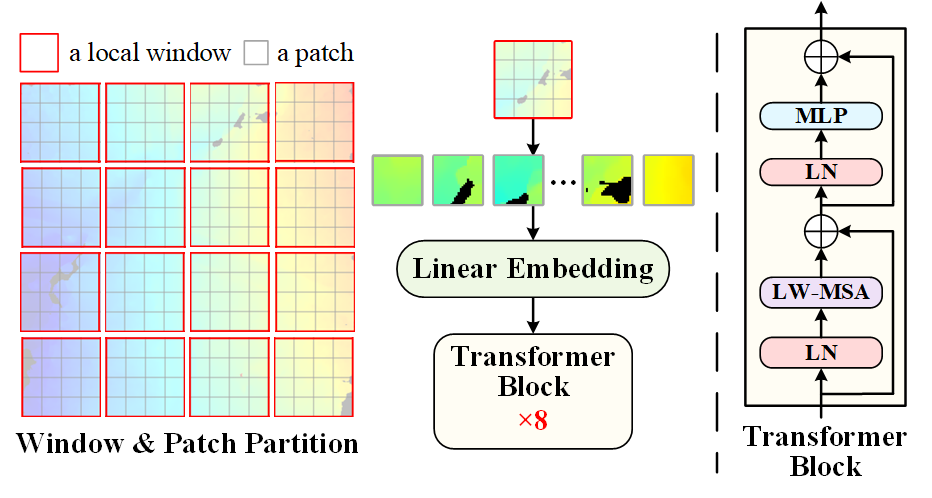} \\
	\caption{An illustration of the local window based transformer encoder. The input terrain feature map is first partitioned into a sequence of local windows, and each local window is further split into several terrain patches. Then, a linear embedding layer is applied on each input patch, followed by eight transformer blocks. The right figure is the compositions of each transformer block. }
	\label{fig_encoder}
	\vspace{-0.6cm} 
\end{figure}
For traversability analysis, it is helpful to effectively utilize terrain information from spatially adjacent terrain patches since traversability itself is a spatially-related concept. However, it is challenging to determine the optimal range of supported spatial neighborhoods. To address this issue and extract more representative embeddings, a local window based transformer encoder $\bm{f}$ inspired by the Swin Transformer \cite{Liu2021} is introduced in this subsection. 

An illustration of the proposed encoder $\bm{f}$ is shown in Fig. \ref{fig_encoder}. The input terrain feature map $\bm{F}$ is partitioned into a sequence of non-overlapping local windows, and each local window is further split into $W \times W$ terrain patches. All terrain patches are automatically labeled using the approach introduced in Section \ref{Section: Label}. A local window is treated as a whole and fed into the $\bm{f}$. Each flatten terrain patch contained in the local window is treated as a token. In $\bm{f}$, a linear embedding layer is applied to the input flatten tokens to project them to a $D$-dimensional embedding $\bm{E} \in \mathbb{R}^{W^2 \times D}$. $\bm{E}$ is then fed into eight transformer blocks. The shape of the output embedding for each transformer block remains unchanged. Each transformer block consists of a local window based multi-head self-attention (LW-MSA) module, followed by a 2-layer MLP module. A Layer Normalization (LN) layer is applied before each LW-MSA and MLP module, and a residual connection is applied after each LW-MSA and MLP module.  The whole process of the proposed encoder $\bm{f}$ can be formulated as:
\begin{equation}
\begin{split}
\bm{Z}^0 &= \bm{E} \,, \\
\bm{\hat{Z}}^l &= \mathrm{\textbf{LW\mbox{-}MSA}}\left(\mathrm{\textbf{LN}}\left(\bm{Z}^{l-1}\right)\right) + \bm{Z}^{l-1} \,, \\
\bm{Z}^l &= \mathrm{\textbf{MLP}}\left(\mathrm{\textbf{LN}}\left(\bm{\hat{Z}}^l\right)\right) + \bm{\hat{Z}}^l \,,
\end{split}
\end{equation}
where $\bm{\hat{Z}}^l \in \mathbb{R}^{W^2 \times D}$ and $\bm{Z}^l \in \mathbb{R}^{W^2 \times D}$ represent the output embedding vectors of the LW-MSA module and the MLP module in the $l$-th transformer block, respectively. The final extracted embedding for each terrain patch $\bm{x}_i$ is $\bm{z}^8_i \in \mathbb{R}^D$ , which is simply denoted as $\bm{z}_i$ in the subsequent contents.  

\subsection{Prototype-based Contrastive Representation Learning}
\begin{figure}[t]
	\centering
	\includegraphics[scale=0.33]{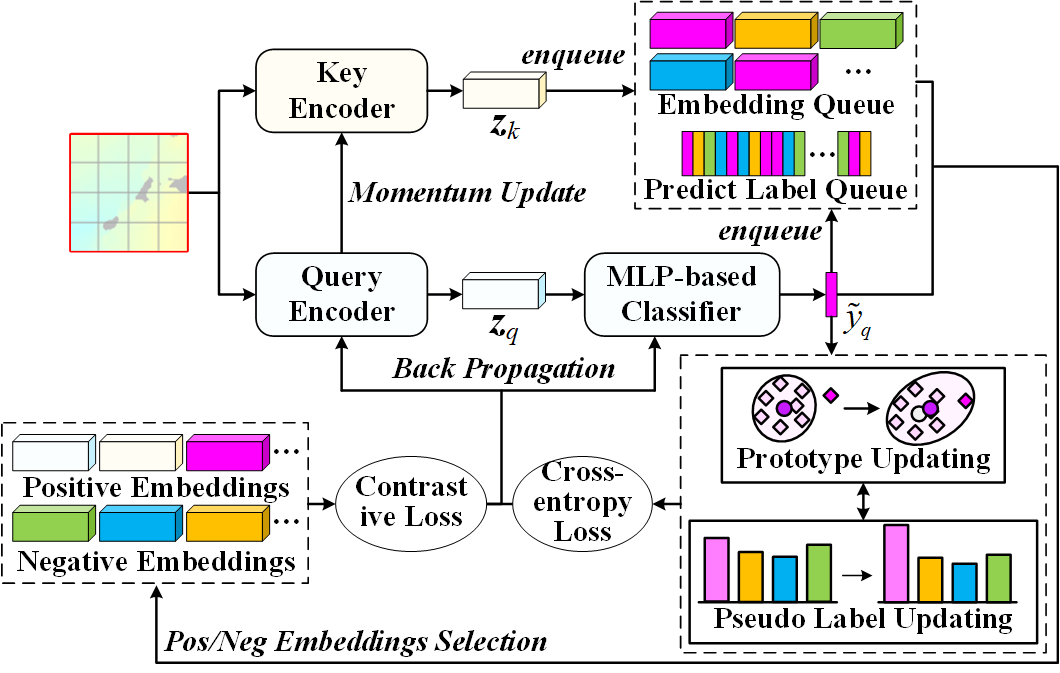} \\
	\caption{The overall process of the prototype-based contrastive representation learning approach.}
	\label{fig_contrastive learning}
	\vspace{-0.6cm} 
\end{figure}
Inspired by \cite{Wang2022}, a prototype-based contrastive representation learning approach is proposed to learn discriminative embeddings for self-supervised traversability learning. The overall process of this approach is illustrated in Fig. \ref{fig_contrastive learning}.

In this approach, a given local window is first processed by separate query encoder $\bm{f}_q$ and key encoder $\bm{f}_k$, respectively, generating a query embedding $\bm{z}_q$ and a key embedding $\bm{z}_k$ for each token. Only the parameters $\bm{\theta}_q$ of $\bm{f}_q$ are updated by back-propagation, while the parameters $\bm{\theta}_k$ of $\bm{f}_k$ are momentum updated by $\bm{\theta}_q$:
\begin{equation}
\bm{\theta}_k = m_{\theta} \cdot \bm{\theta}_k + \left(1-m_{\theta}\right) \cdot \bm{\theta}_q \,,
\end{equation}
where $m_{\theta}$ is a momentum coefficient for updating encoder. The query embedding $\bm{z}_q$ is then fed into an MLP-based classifier $\bm{f}_c$. By combining the output of $\bm{f}_c$ with the pseudo label $\bm{y}$ of the token corresponding to $\bm{z}_q$, a masked predicted label $\tilde{y}_q$ can be generated by:
\begin{equation}
\tilde{y}_q = \mathop{\arg\max}\limits_{j \in \left[1, K\right]} \left[\bm{f}_c^j\left(\bm{z}_q\right) \cdot \bm{y}\right] \,,
\end{equation}
where $\bm{f}_c^j\left(\bm{z}_q\right)$ denotes the $j$-th component of the output vector $\bm{f}_c\left(\bm{z}_q\right) \in \mathbb{R}^K$. 

An embedding queue $\bm{Q}_e$ and a predicted label queue $\bm{Q}_l$ are maintained to store the recently encoded key embeddings and their corresponding predicted labels. The embeddings and labels of the latest tokens are enqueued, and the same number of the oldest embeddings and labels are dequeued to ensure a fixed queue size. For a token $\bm{x}$ with a predicted label $\tilde{y}_q$, its positive embeddings can be selected from $\bm{Q}_e$. Specifically, any embedding $\bm{z}'$ in $\bm{Q}_e$ with the same predicted label as $\tilde{y}_q$ is selected as a positive embedding, while the remaining embeddings are considered as negative embeddings. After the positive/negative embeddings selection, the per-token contrastive loss $\mathcal{L}_{\mathrm{cont}} \left(\bm{x}\right)$ can be defined as:
\begin{equation}
\mathcal{L}_{\mathrm{cont}} \left(\bm{x}\right) = \frac{-1}{\left|\bm{A}\left(\bm{x}\right)\right|} \! \sum_{\bm{z}^{+} \in \bm{A}\left(\bm{x}\right)} \!\! \log \frac{\exp \left(\bm{z}_q^T \! \cdot \! \bm{z}^{+} / \tau \right)}{\sum\limits_{\bm{z}_j \in \bm{Q}_e} \! \! \exp \left(\bm{z}_q^T \! \cdot \! \bm{z}_j/\tau\right)} \,,
\end{equation}
where $\tau$ is a temperature hyper-parameter, and $\left|\bm{A}\left(\bm{x}\right)\right|$ denotes the total number of positive samples in the positive embedding set $\bm{A}\left(\bm{x}\right)$.

To ensure the generation of discriminative embeddings, a high-quality classifier is required for accurate positive/negative embeddings selection. However, improving the performance of the classifier solely through the contrastive loss is challenging due to the inherent ambiguity of pseudo labels. To alleviate this problem, $K$ prototype vectors $\bm{\Psi} = \left\{\bm{\psi}_c\right\}_{c=1:K}$ are created for incremental updating of the pseudo labels. Each prototype serves as a representative embedding for a group of similar embeddings. During training, $\bm{\psi}_c$ is momentum updated by those query embeddings $\bm{z}_q$ whose predicted labels $\tilde{y}_q$ belong to class $c$, and the update process can be expressed as: 
\begin{equation}
\bm{\psi}_c = \frac{m_p \cdot \bm{\psi}_c + \left(1-m_p\right) \cdot \bm{z}_q}{\Vert m_p \cdot \bm{\psi}_c + \left(1-m_p\right) \cdot \bm{z}_q\Vert_2} \,, 
\end{equation}
where $m_{p}$ is a momentum coefficient for updating the prototypes, $\Vert \cdot \Vert_2$ denotes L2-norm of a vector. 

After the prototype updating, the pseudo label updating process is performed. An initial normalized vector $\bm{y}_n = \frac{1}{\sum_{i=1}^{K} \bm{y}} \bm{y}$ is assigned to each token based on its pseudo label $\bm{y}$ in the first batch. Then, an indicator vector $\bm{\xi} \in \mathbb{R}^K$ is computed by comparing the similarity between $\bm{z}_q$ and $\bm{\Psi}$, and $\bm{y}_n$ is momentum updated by:  
\begin{align}
\bm{y}_n &= m_l \cdot \bm{y}_n + \left(1 - m_l\right) \cdot \bm{\xi}\,,  \\
\bm{\xi}^c &= \left\{ 
\begin{matrix}
\begin{aligned}
& 1 && \mathrm{if} \,\, c = \mathop{\arg\max}\limits_{j \in \left[1, K\right]} \left(\bm{z}_q^T \cdot \bm{\psi}_j\right) \\
& 0 && \mathrm{else} 
\end{aligned}
\end{matrix} \right. \,, 
\end{align}
where $m_{l}$ is a momentum coefficient for updating pseudo labels, and $\bm{\xi}^c$ denotes the $c$-th component of $\bm{\xi}$. $\bm{y}_n$ is considered as the refined pseudo label, and is utilized for calculating the per-token cross-entropy loss $\mathcal{L}_{\mathrm{cls}} \left(\bm{x}\right)$ as:
\begin{equation}
\mathcal{L}_{\mathrm{cls}} \left(\bm{x}\right) = \sum_{j=1}^{K} -\bm{y}_n^j \cdot \log\left(\bm{f}_c^j \left(\bm{z}_q\right)\right)\,.
\end{equation}

In the training process, the MLP-based classifier and the query encoder are jointly trained, and the overall loss function is:
\begin{equation}
\mathcal{L}_{\mathrm{sum}} = \mathcal{L}_{\mathrm{cls}} + \lambda \mathcal{L}_{\mathrm{cont} }\,,
\label{equation_loss}
\end{equation}
where $\lambda$ is a weight used for balancing $\mathcal{L}_{\mathrm{cls}}$ and $\mathcal{L}_{\mathrm{cont}}$. 

In summary, the proposed prototype-based contrastive representation learning approach consists of two components that mutually reinforce each other. The discriminative embeddings learned from contrastive learning enhance the quality of positive/negative embeddings selection, while the refined pseudo labels in turn improve the performance of contrastive representation learning. As the iterative interaction of prototype updating and pseudo label updating, the ambiguities associated with those pseudo labels are gradually eliminated, leading to the understanding of the specific traversability. 

\section{EXPERIMENTAL RESULTS} \label{Section: experiment}
\subsection{Experimental Datasets}
\begin{figure}[t]
	\centering
	\includegraphics[scale=0.32]{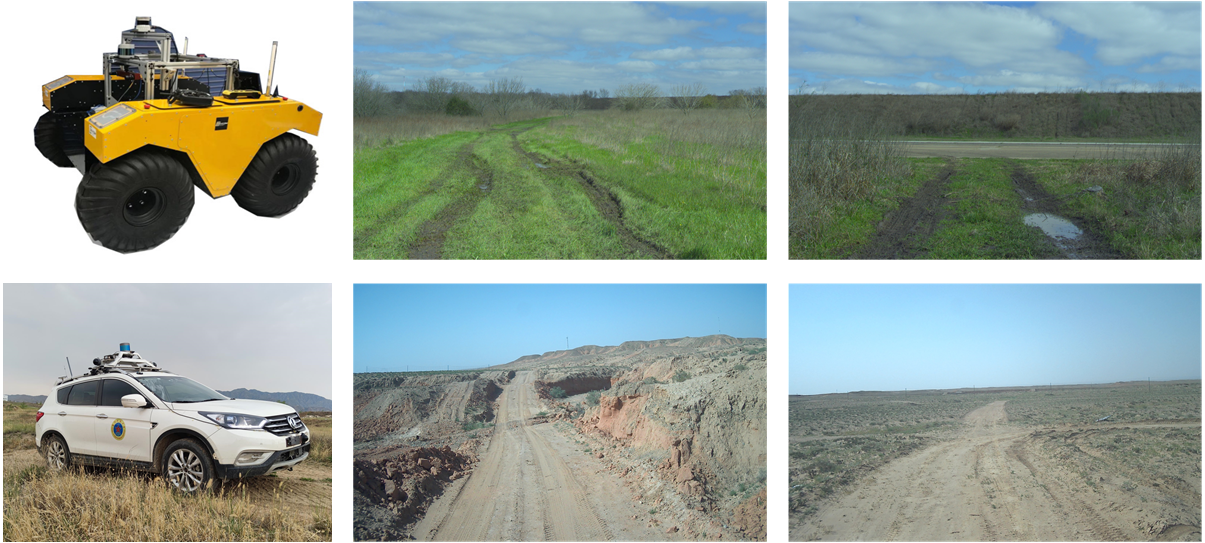} \\
	\caption{Data collection platforms and typical scenes of RELLIS-3D dataset (the top figures) and our Gobi Desert driving dataset (the bottom figures).}
	\label{fig_dataset}
	\vspace{-0.6cm} 
\end{figure}

To evaluate the proposed method, experiments are conducted on two off-road datasets: the publicly available RELLIS-3D \cite{Jiang2021} dataset and a Gobi Desert driving dataset collected by our UGV. The data collection platforms and some typical scenes of both datasets are shown in Fig. \ref{fig_dataset}.

The RELLIS-3D dataset consists of five sequences of LiDAR frames collected in a rugged off-road environment using a Warthog all-terrain UGV. The UGV is equipped with an Ouster OS1 LiDAR and a Vectornav VN-300 inertial navigation system. Each LiDAR frame is point-wise annotated with 20 different semantic classes (such as grass, fence, tree, barrier, etc.). Additionally, ground-truth pose for each frame is provided by a high-precision Simultaneous Localization and Mapping (SLAM) system. For our experiments, we select 50 key-frames from sequence 01 for training, 200 random frames from the remaining frames of sequence 01 for validation, and all 2059 frames from sequence 04 for quantitative and qualitative testing.

In our Gobi Desert driving dataset, LiDAR frames were collected in a Gobi desert scene. Our UGV is equipped with a Robosense RS-Ruby128 LiDAR and a StarNeto XW-GI7660 GNSS/INS system. High-frequency 6-degree of freedom (DoF) poses with centimeter-level accuracy can be obtained by using an online pose estimation module proposed in our previous work \cite{Xue2019}. For our experiments, we select 100 key-frames for training, and 1900 frames for qualitative testing.

\subsection{Evaluation Metrics}
To quantitatively evaluate the performance of the proposed method, we utilize the annotations from the RELLIS-3D dataset to generate ground-truth traversability maps. The semantic categories are grouped into three traversability levels (traversable, non-traversable, and risky) based on their travel costs.
In the process of ground-truth generation, several annotated LiDAR frames are first assembled by using the provided ground-truth poses. The merged dense point cloud is then projected onto a 2D grid map, and the traversability of each grid cell is determined by the semantic labels of the projected points. If all the projected points within a grid cell have labels such as ``grass", ``puddle", ``asphalt", or ``concrete", it is considered as a traversable cell; if the labels of all projected points are ``bush" or ``fence", it is considered as a risky cell; otherwise, it is considered as a non-traversable cell.

Given the ground-truth traversability maps, we evaluate the traversability analysis results by two performance metrics widely used for semantic segmentation: Pixel Accuracy (PA) and mean Intersection over Union (mIoU). PA measures the proportion of correctly classified grid cells in the prediction results, and mIoU calculates the degree of overlap between the ground-truth and prediction results. Both metrics provide a quantitative measure of the grid-level prediction accuracy.

\subsection{Implementation Details}
In our experiments, we set the resolution of each grid cell to $0.2\mathrm{m} \times 0.2\mathrm{m}$, and the map size is set to $40\mathrm{m} \times 40\mathrm{m}$. Each terrain patch consists of $11 \times 11$ ($M = 11$) grid cells, and each local window comprises $10 \times 10$ ($W = 10$) terrain patches. The dimensionality $D$ of the embedding is set to 32. The lengths of the embedding queue $\bm{Q}_e$ and the predicted label queue $\bm{Q}_l$ are kept as 81920. The momentum coefficients $m_{\theta}$ and $m_{p}$ are set to 0.999 and 0.99, respectively. The initial momentum coefficient $m_{l}$ is set to 0.99, and its value decays polynomially after the initial 10 epochs. For hyper-parameters, we set $\tau$ to 0.07, and $\lambda$ to 0.5. During training, we use Stochastic Gradient Descent (SGD) as the optimizer, with a weight decay of $1e^{-5}$, a momentum of 0.9, and an initial learning rate of 0.02. The network is trained for 50 epochs on a NVIDIA RTX A6000 GPU, with an exponentially decayed learning rate. 

\subsection{Ablation Studies}
\subsubsection{Prototype Num}
\begin{figure}[t]
	\centering
	\includegraphics[scale=0.23]{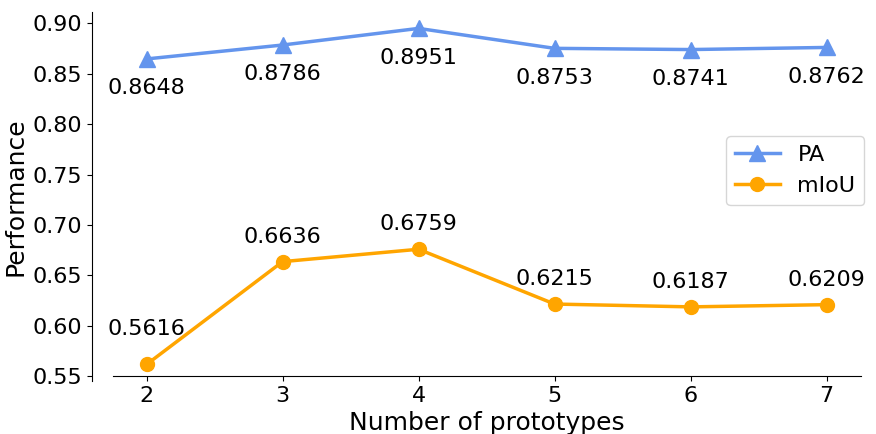} \\
	\caption{Performance comparison results of the ablation study with varying number of prototypes $K$.}
	\label{fig_ablation_Prototypes}
	\vspace{-0.2cm}
\end{figure}

To evaluate how different number of prototypes $K$ affects the performance of the proposed method, an ablation study is conducted with varying number of $K$. The quantitative experimental results are presented in Fig \ref{fig_ablation_Prototypes}. It can be found that an increasing number of $K$ boosts the model's performance until when $K = 4$, and then the model's performance decreases and tends to converge when $K > 4$. Therefore, we choose $K = 4$ as the optimal number of prototypes for our subsequent experiments.

\begin{figure}[t]
	\centering
	\includegraphics[scale=0.44]{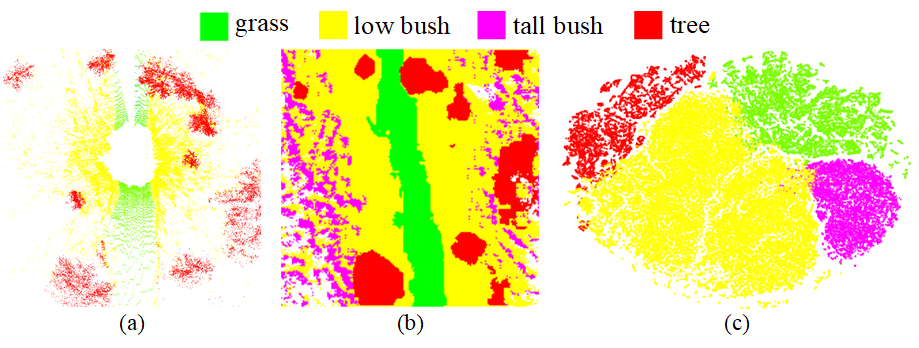} \\
	\caption{(a) visualizes the semantic annotation information of a raw LiDAR scan. (b) shows the traversability classification result. (c) is the 2D t-SNE visualization of the generated embeddings. Different colors represent different semantic categories. }
	\label{fig_pototypes_analysis}
	\vspace{-0.6cm}
\end{figure}

Furthermore, we also conduct visualization to gain insights into the generated prototypes and their semantic meaning. First, we visualize the traversability classification results (Fig.\ref{fig_pototypes_analysis}(b)). Notably, the proposed method automatically divides the ``bushes" category into ``tall bushes" and ``low bushes", resulting a finer semantic categories compared to the original annotations (Fig.\ref{fig_pototypes_analysis}(a)). Subsequently, we employ t-SNE \cite{Maaten2008} visualization to explore the embedding space (Fig.\ref{fig_pototypes_analysis}(c)). We observe that well-separated clusters are generated in the embedding space. Each cluster represents a specific semantic category, and can be represented by a prototype. Based on these visualizations, we can interpret the semantic meaning of each prototype. In the subsequent traversability analysis, the grass category corresponds to traversable regions, the tree category corresponds to non-traversable regions, and both low bushes and high bushes are considered as risk regions.

\subsubsection{Input data}
To verify the validity of the input terrain feature map $\bm{F}$ in the proposed method, we conduct an ablation study by varying the forms of input data. In LiDAR-based traversability analysis approaches, a common input data format is the BEV grid map \cite{Gao2019, Suger2015}. In this ablation study, we consider two common variations of BEV grid maps: the single LiDAR scan BEV (S-BEV) and the multiple LiDAR scans BEV (M-BEV). The S-BEV is generated from a single LiDAR scan, while the M-BEV is formed by fusing multiple LiDAR scans. The S-BEV and M-BEV are applied as two forms of input data in the proposed framework for comparative analysis. The quantitative experimental results are shown in Table \ref{tabel_ablation_study}. The results indicate that the model achieves the worst performance when using S-BEV as the input data. This can be attributed to the sparse nature of a single LiDAR scan, which may fail to provide stable and complete representations of the local environment. Although the model's performance improves significantly when using M-BEV as the input data, there still exists a performance gap compared to using $\bm{F}$ as the input data. The reason is that $\bm{F}$ contains richer information compared to M-BEV.

\begin{table}[t]
	\centering
	\setlength{\abovecaptionskip}{-0.01cm}
	\caption{Performance results of different ablation studies.}  
	\begin{tabular}{ccccc} 
		\toprule
		\textbf{Input} & \textbf{Encoder}  & \textbf{Loss} & \textbf{PA}& \textbf{mIoU} \\
		\midrule
		S-BEV & \multirow{2}{*}{LW-Transformer} & \multirow{2}{*}{$\mathcal{L}_{\mathrm{sum}}$} & 0.6477 & 0.3816 \\
		M-BEV &  &  & 0.8679 & 0.6469 \\
		\midrule
		\multirow{2}{*}{$\bm{F}$} & ResNet-18 & \multirow{2}{*}{$\mathcal{L}_{\mathrm{sum}}$} & 0.7939 & 0.5785  \\
		& AlexNet & &  0.8161 & 0.6071 \\
		\midrule
		\multirow{2}{*}{$\bm{F}$} & \multirow{2}{*}{LW-Transformer} & $\mathcal{L}_{\mathrm{cont}}$ & 0.7371 & 0.5428 \\
		&  &  $\mathcal{L}_{\mathrm{cls}}$ & 0.6861 & 0.4531 \\
		\midrule
		$\bm{F}$  & LW-Transformer & $\mathcal{L}_{\mathrm{sum}}$ & \textbf{0.8951} & \textbf{0.6759} \\
		\bottomrule	
	\end{tabular}
	\label{tabel_ablation_study}
	\vspace{-0.6cm}
\end{table}
\subsubsection{Encoder Network}
To evaluate the validity of the proposed local window based transformer (LW-Transformer), two commonly used backbone networks (AlexNet and ResNet-18) are employed as encoders in the proposed framework for comparative analysis. The results in Table \ref{tabel_ablation_study} show that model's performance decreases when using AlexNet or ResNet-18 as encoders. The primary reason is that traversability is a spatially-related concept, the traversability of a terrain patch not only depends on the patch itself, but also on those neighboring terrain patches within a certain range. The self-attention mechanism incorporated in the LW-Transformer enables it to capture the implicit spatial dependencies between adjacent terrain patches. This capability is crucial for accurate traversability analysis. In contrast, CNN-based encoder networks lack the modeling of spatial dependencies,  which results in performance degradation.

\begin{table}[b]
	\vspace{-0.4cm}
	\centering
	\setlength{\abovecaptionskip}{-0.01cm}
	\caption{Performance results of different methods.}  
	\renewcommand{\arraystretch}{1.2}
	\begin{tabular}{ccc} 
		\toprule
		\textbf{Approaches} & \textbf{PA} & \textbf{mIoU}\\
		\midrule
		Rule-based \cite{Xue2023}  & 0.8239 & 0.5382 \\	
		ORDAE-Net \cite{Gao2019} & 0.5053 & 0.3279 \\
		Ours & \textbf{0.8951} & \textbf{0.6759} \\
		\bottomrule	
	\end{tabular}
	\label{tabel_comparison}
\end{table}

\subsubsection{Loss Function}
To validate the impact of the loss functions $\mathcal{L}_{\mathrm{cont}}$ and $\mathcal{L}_{\mathrm{cls}}$ in the prototype-based contrastive representation learning, we conduct an ablation study by considering each loss function individually. The results presented in Table \ref{tabel_ablation_study} clearly indicate that the model's performance decreases significantly when utilizing only $\mathcal{L}_{\mathrm{cont}}$ or $\mathcal{L}_{\mathrm{cls}}$ as the loss function. This finding validates the necessity of using a joint loss function that combines $\mathcal{L}_{\mathrm{cont}}$ and $\mathcal{L}_{\mathrm{cls}}$ as Eq. (\ref{equation_loss}) in the proposed method.

\subsection{Comparative Experiments}
\begin{figure}[t]
	\centering
	\subfigure {
		\begin{minipage}[t]{0.205\linewidth}
			\centering
			\includegraphics[scale=0.33]{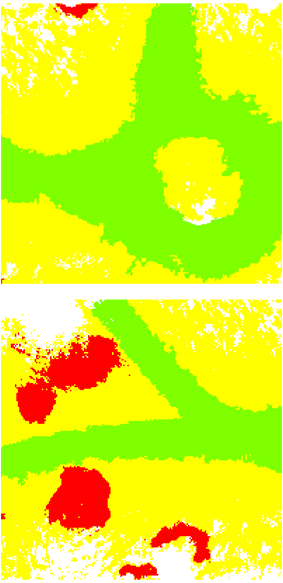}
			\centerline{\,\, \footnotesize ({a}) Ground-truth}
			\label{fig_comparative_a}
		\end{minipage}
	} %
	\hfill
	\subfigure{
		\begin{minipage}[t]{0.205\linewidth}
			\centering
			\includegraphics[scale=0.33]{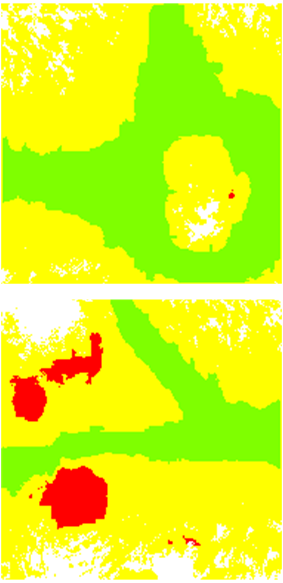}
			\centerline{\footnotesize ({b}) Ours}
			\label{fig_comparative_b}
		\end{minipage}
	} %
	\hfill
	\subfigure{
		\begin{minipage}[t]{0.205\linewidth}
			\centering
			\includegraphics[scale=0.33]{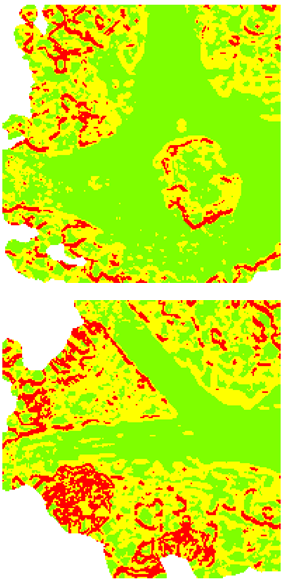}
			\centerline{\footnotesize ({c}) Rule-based}
			\label{fig_comparative_c}
		\end{minipage}
	} %
	\hfill
	\subfigure{
		\begin{minipage}[t]{0.205\linewidth}
			\centering
			\includegraphics[scale=0.33]{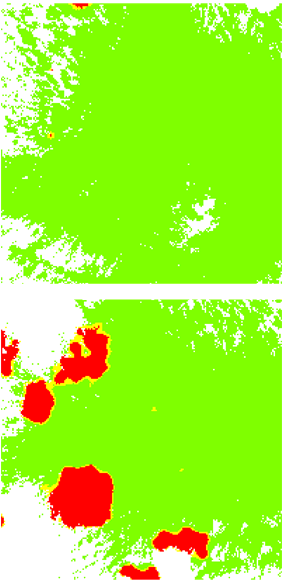}
			\centerline{\footnotesize ({d}) ORDAE-Net}
			\label{fig_comparative_d}
		\end{minipage}
	} 
	\setlength{\abovecaptionskip}{-0.3cm}
	\caption{Qualitative comparison results on RELLIS-3D dataset. (a) is the ground-truth, (b)-(d) are traversability analysis results generated by the proposed method, the rule-based approach \cite{Xue2023}, and the ORDAE-Net \cite{Gao2019}, respectively. The green cells indicate traversable regions, the red cells represent non-traversable regions, and the yellow cells is risky regions.}
	\label{fig_comparative}
\end{figure}

We compare the proposed method with two recent LiDAR-based traversability analysis approaches. The first one \cite{Xue2023} is a rule-based approach that estimates the travel cost for each region by using the constructed 3D terrain models. It determines traversability based on some cost thresholds derived from vehicle trajectories. The second approach is a self-supervised learning based off-road drivable area extraction network (ORDAE-Net) \cite{Gao2019}. ORDAE-Net segments the environments into obstacle regions, traversable regions, and grey regions using vehicle paths and auto-generated obstacle labels. Fig. \ref{fig_comparative} shows some qualitative
comparison results, and the quantitative results are presented in Table \ref{tabel_comparison}. The results in Fig. \ref{fig_comparative} show that the rule-based approach can roughly distinguish the overall shape of regions with different traversability, but the results often contain a significant amount of noise. The ORDAE-Net excels in detecting non-traversable regions but struggles to distinguish traversable regions from those similar risky regions. In contrast, the proposed method demonstrates superior capability in distinguishing regions with varying traversability. The qualitative results shown in Table \ref{tabel_comparison} further supports the superiority of the proposed method. It significantly surpasses the other two approaches in terms of both PA and mIoU. 

\subsection{Qualitative Results on Gobi Desert Driving Dataset}
\begin{figure}[t]
	\centering
	\includegraphics[scale=0.36]{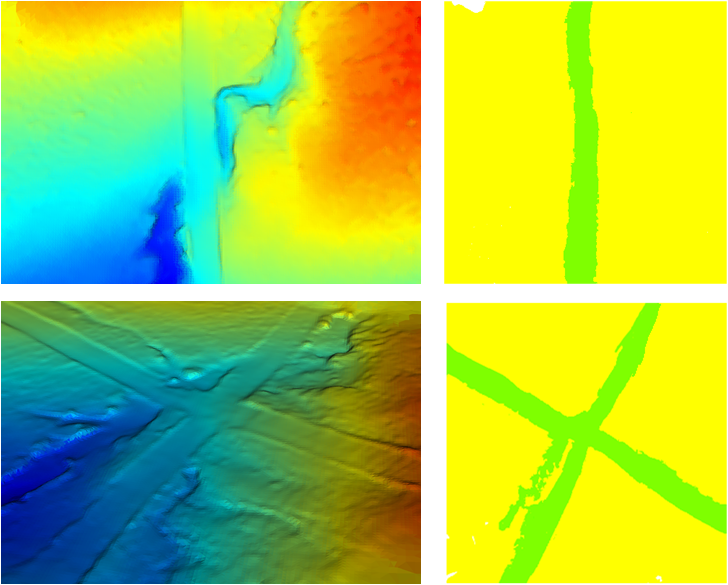} \\
	\caption{The qualitative results on the Gobi Desert driving dataset. Left figures show the terrain models, and right figures is the corresponding traversability analysis results generated by the proposed method.}
	\label{fig_ALS_results}
	\vspace{-0.6cm}
\end{figure}
We further conduct experiments on our Gobi Desert driving dataset to evaluate whether the proposed method can be adapted to different off-road environments. Some qualitative results are presented in Fig. \ref{fig_ALS_results}. It can be observed that the proposed method works well in the Gobi environment which has a high degree of terrain similarity, and accurately identifies those traversable regions. However, as shown in the bottom figure of Fig. \ref{fig_ALS_results}, some gullies are misclassified as traversable regions since the terrain features of these regions closely resemble those of traversable regions, leading to this incorrect classification.

\section{CONCLUDING REMARKS} \label{Section: Conclusion}
In this paper, we present a novel terrain traversability learning method that leverages a contrastive label disambiguation strategy to learn platform-specific and task-specific traversability in a self-supervised manner, without any human-provided annotations. To achieve this, a prototype-based contrastive representation learning approach is designed to learn discriminative embeddings by using weakly labeled terrain patches obtained from actual driving experiences. As the iterative interaction between prototype updating and pseudo label updating, the ambiguities of those pseudo labels are gradually eliminated, and the specific traversability can be learned. Experimental results on both the RELLIS-3D dataset and our Gobi Desert driving dataset have demonstrated the effectiveness of the proposed method. In future work, we aim to address the limitations of using LiDAR as sole sensing modality by incorporating visual and proprioceptive modalities to capture richer terrain features.

\section*{ACKNOWLEDGMENT}

This work was supported by the National Natural Science Foundation of China under No. 61790565 and No. 61803380.

\end{document}